\pdfoutput=1
\documentclass[11pt]{article}

\usepackage{emnlp2021}

\usepackage{times}
\usepackage{latexsym}

\usepackage[T1]{fontenc}
\usepackage[utf8]{inputenc}

\usepackage{microtype}

\usepackage{amsmath, amsfonts}
\usepackage{multirow}
\usepackage{graphicx, graphics, subfigure}
\usepackage{booktabs}
\usepackage{tablefootnote}
\usepackage{footnote}
\usepackage{soul}
\title{Robust Unsupervised Cross-Lingual Word Embedding \\
using Domain Flow Interpolation}

\author{
  Liping Tang\textsuperscript{1} \quad Zhen Li\textsuperscript{2} \quad Zhiquan Luo\textsuperscript{2} \quad Helen Meng\textsuperscript{1,3}\\
    \textsuperscript{1}Centre for Perceptual and Interactive Intelligence \\
    \textsuperscript{2}The Chinese University of Hong Kong, Shenzhen \\
    \textsuperscript{3}The Chinese University of Hong Kong \\
  \texttt{lptang@cpii.hk, \{zhenli, zqluo\}@cuhk.edu.cn, hmmeng@cuhk.edu.hk} \\
  }

\begin{document}
\maketitle
\begin{abstract}
This paper investigates an unsupervised approach towards deriving a universal, cross-lingual word embedding space, where words with similar semantics from different languages are close to one another. Previous adversarial approaches have shown promising results in inducing cross-lingual word embedding without parallel data. However, the training stage shows instability for distant language pairs. Instead of mapping the source language space directly to the target language space, we propose to make use of a sequence of intermediate spaces for smooth bridging. Each intermediate space may be conceived as a pseudo-language space and is introduced via simple linear interpolation. This approach is modeled after domain flow in computer vision, but with a modified objective function. Experiments on intrinsic Bilingual Dictionary Induction tasks show that the proposed approach can improve the robustness of adversarial models with comparable and even better precision. Further experiments on the downstream task of Cross-Lingual Natural Language Inference show that the proposed model achieves significant performance improvement for distant language pairs in downstream tasks compared to state-of-the-art adversarial and non-adversarial models.
\end{abstract}

\section{Introduction}

Learning cross-lingual word embedding (CLWE) is a fundamental step towards deriving a universal embedding space in which words with similar semantics from different languages are close to one another. CLWE has also shown effectiveness in knowledge transfer between languages for many natural language processing tasks, including Named Entity Recognition \citep{guo2015cross}, Machine Translation \citep{gu2018universal}, and Information Retrieval \citep{vulic2015monolingual}. 

Inspired by \citet{mikolov2013exploiting}, recent CLWE models have been dominated by \textit{mapping-based} methods (\citealp{ruder2019survey}; \citealp{glavas2019properly}; \citealp{vulic2019we}). They map monolingual word embeddings into a shared space via linear mappings, assuming that different word embedding spaces are nearly isomorphic.
By leveraging a seed dictionary of 5000 word pairs, \citet{mikolov2013exploiting} induces CLWEs by solving a least-squares problem. Subsequent works (\citealp{xing2015normalized}; \citealp{artetxe2016learning}; \citealp{smith2017offline}; \citealp{joulin2018loss}) propose to improve the model by normalizing the embedding vectors, imposing an orthogonality constraint on the linear mapping, and modifying the objective function.
Following work has shown that reliable projections can be learned from weak supervision by utilizing shared numerals\cite{artetxe2017learning}, cognates \cite{smith2017offline}, or identical strings \cite{sogaard2018limitations}.
 
Moreover, several fully unsupervised approaches have been recently proposed to induce CLWEs by adversarial training (\citealp{zhang2017adversarial}; \citealp{zhang2017earth}; \citealp{conneau2017word}). 
State-of-the-art unsupervised adversarial approaches (\citealp{conneau2017word}) have achieved very promising results and even outperform supervised approaches in some cases. 
However, the main drawback of adversarial approaches lies in their instability on distant language pairs \cite{sogaard2018limitations}, inspiring the proposition of non-adversarial approaches (\citealp{hoshen2018non}; \citealp{artetxe2018robust}). In particular,  \citet{artetxe2018robust} (VecMap) have shown strong robustness on several language pairs. However, it still fails on 87 out of 210 distant language pairs \citep{vulic2019we}. 

Subsequently, \citet{li2020simple} proposed Iterative Dimension Reduction to improve the robustness of VecMap. On the other hand, \citet{mohiuddin2019revisiting} revisited adversarial models and add two regularization terms that yield improved results. However, the problem of instability still remains. For instance, our experiments show that the improved version\citep{mohiuddin2019revisiting} still fails in inducing reliable English-to-Japanese and English-to-Chinese CLWE space.
% \footnote{The code of \citet{li2020simple} is not available and thus }
% , especially for distant language pairs. 

In this paper, we focus on the challenging task of unsupervised CLWE on distant language pairs.
Due to the high precision achieved by adversarial models, we revisit adversarial models and  propose to improve their robustness.
We adopt the network architecture from \citet{mohiuddin2019revisiting} but treat the unsupervised CLWE task as a domain adaptation problem. 
Our approach is inspired by the idea of domain flow in computer vision that has been shown to be effective for domain adaptation tasks. 
\citet{gong2019dlow} introduced intermediate domains to generate images of intermediate styles. They added an intermediate domain variable on the input of the generator via conditional instance normalization. The intermediate domains can smoothly bridge the gap between source and target domains to ease the domain adaptation task.
Inspired by this idea, we adapt domain flow for our task by introducing intermediate domains via simple linear interpolations. Specifically, rather than mapping the source language space directly to the target language space, we map the source language space to intermediate ones. Each intermediate space may be conceived as a pseudo-language space and is introduced as a linear interpolation of the source and target language space. 
We then engage the intermediate space approaching the target language space gradually. Consequently, the gap between the source language space and the target space could be smoothly bridged by the sequence of intermediate spaces.
% We make use of a sequence of such interpolated intermediate spaces to smoothly bridge the gap between the source language space and the target space.
We have also modified the objective functions of the original domain flow for our task.
% We then engage the intermediate space approaching the target language space gradually. Consequently, the gap between the source language space and the target space could be smoothly bridged by the sequence of intermediate spaces.
% Additionally, the objective functions of the original domain flow are modified for our task. 

We evaluate the proposed model on both intrinsic and downstream tasks. Experiments on intrinsic Bilingual Dictionary Induction (BLI) tasks show that our method can significantly improve the robustness of adversarial models.
Simultaneously, it could achieve comparable or even better precision compared with the state-of-the-art adversarial and non-adversarial models. 
Although BLI is a standard evaluation task for CLWEs, the performance on the BDI task might not correlate with performance in downstream tasks\citep{glavas2019properly}.
Following previous works (\citealp{glavas2019properly}; \citealp{doval2019robustness}; \citealp{ormazabal2020beyond}), we choose Cross-Lingual Natural Language Inference (XNLI), a language understanding task, as the downstream task to further evaluate the proposed model.
Experiments on the XNLI task show that the proposed model achieves higher accuracy on distant language pairs compared to baselines, which validates the importance of robustness of CLWE models in downstream tasks and demonstrates the effectiveness of the proposed model.

\section{Proposed Model}
\label{sec:model}

Our model is implemented based on the network structure from \citet{mohiuddin2019revisiting}, which implements a cycleGAN on the latent word representations transformed by autoencoders.
In our model, the source language space corresponds to the source domain $\mathcal{S}$ and the target language space corresponds to the target domain $\mathcal{T}$. 

\subsection{Introducing Intermediate Domains}
\label{sec:interdom}

Let $z\in[0,1]$ and denote the intermediate domain as $\mathcal{M}^{(z)}$, similar to \citet{gong2019dlow}. $\mathcal{M}^{(0)}$ corresponds to the source domain $\mathcal{S}$, and $\mathcal{M}^{(1)}$ corresponds to the target domain $\mathcal{T}$. By varying $z$ from 0 to 1, we can obtain a sequence of intermediate domains from $\mathcal{S}$ to $\mathcal{T}$, referred to as domain flow.
There are many possible paths from $\mathcal{S}$ to $\mathcal{T}$ and we expect $\mathcal{M}^{(z)}$ to be the shortest one.

Moreover, given any z, we expect the distance between $\mathcal{S}$ and $\mathcal{M}^{(z)}$ to be proportional to the distance between $\mathcal{S}$ and $\mathcal{T}$ by z, 
% We consider the distance between domains as the distance between the data distributions of domains. 
% Denote the data distributions for the source domain $\mathcal{S}$, the target domain $\mathcal{T}$ and the intermediate domain $\mathcal{M}^{(z)}$ as $P_S$, $P_T$ and $P_{M^{(z)}}$, respectively, then we expect
% \begin{equation}
% \frac{\operatorname{dist}\left(P_{S}, P_{M^{(z)}}\right)}{\operatorname{dist}\left(P_{S}, P_T\right)}=z,
% \end{equation}
or equivalently,
\begin{equation}
\frac{\operatorname{dist}\left(P_{S}, P_{M^{(z)}}\right)}{\operatorname{dist}\left(P_{T}, P_{M^{(z)}}\right)}=\frac{z}{1-z}.
\end{equation}
% Then we have the following equation:
% \begin{equation}
% (1-z) \cdot \operatorname{dist}\left(P_{S}, P_{M^{(z)}}\right)=z \cdot \operatorname{dist}\left(P_{T}, P_{M^{(z)}}\right).
% \end{equation}
 
Thus finding the shortest path from $\mathcal{S}$ to $\mathcal{T}$, i.e., the sequence of $\mathcal{M}^{(z)}$, leads to minimizing the following loss:
\begin{equation}
\mathcal{L}= z \cdot \operatorname{dist}\left(P_{T}, P_{M^{(z)}}\right) + (1-z) \cdot \operatorname{dist}\left(P_{S}, P_{M^{(z)}}\right).
\label{eq:loss1}
\end{equation}

% There are many choices for the distance measure of distributions. The adversarial loss in GAN \citep{goodfellow2014generative} could be viewed as the JS divergence between two distributions (plus a constant) when the discriminator is optimal. In our model, we use the adversarial loss to model the distance between distributions, just like what \cite{gong2019dlow} do.

We use the adversarial loss in GAN \citep{goodfellow2014generative} to model the distance between distributions, similar to \citet{gong2019dlow}.

\subsection{Implementation of Generators}
\label{sec:gen}

Suppose \textbf{x} is sampled from the source domain $\mathcal{S}$ and \textbf{y} is sampled from the target domain $\mathcal{T}$.

The generator $G_{ST}$ in our model transfers data from the source domain to an intermediate domain instead of the target domain. Denote $\mathcal{Z} = [0,1]$, then $G_{ST}$ is a mapping from $ \mathcal{S} \times \mathcal{Z}$ to $\mathcal{M}^{(z)}$.

To ensure the generator to be a linear transformation, we consider our generator as
\begin{equation}
    G_{ST} (\mathbf{x}, z) = \mathbf{W}_{ST}(z)\cdot \mathbf{x} + (1-z) \cdot \mathbf{x}.
\end{equation} 
In this setup, $G_{ST} (\mathbf{x}, 0) = \mathbf{x}$ and $G_{ST} (\mathbf{x}, 1) = \mathbf{W}_{ST}(z)\cdot \mathbf{x}$.
% In this setup, when $z=0$, $G_{ST} (\mathbf{z}_x, z) = \mathbf{z}_x$ and when $z=1$, $G_{ST} (\mathbf{z}_x, z) = \mathbf{W}_{ST}(z)\cdot \mathbf{z}_x$.
We adopt $\mathbf{W}_{ST}(z)$ as a simple scale multiplication on a matrix, i.e., 
\begin{equation}
    \mathbf{W}_{ST}(z) = z\cdot \mathbf{W}_{ST},
\end{equation}
where $\mathbf{W}_{ST}$ is the final transformation matrix that we are interested in. 
% This simple scale multiplication is equivalent to a fully-connected layer without nonlinear activation function where the number of input neurons is 1 and the number of output neurons is equivalent to the dimension of the linear transformation matrix. 
Finally, our intermediate mappings become
\begin{equation}
    G_{ST}(\mathbf{x},z) = z\cdot \mathbf{W}_{ST}\cdot\mathbf{x} + (1-z)\cdot \mathbf{x}.
\end{equation}
These intermediate mappings are simple linear interpolations between the data from source domain $\mathbf{x}$ and that from the pseudo target domain $\mathbf{W}_{ST}\cdot\mathbf{x}$. The generator $G_{TS}(\mathbf{y},z)$ can be defined similarly.

\subsection{The Domain Flow Model}
\label{sec:dlow}
The discriminator $D_S$ is used to distinguish $\mathcal{S}$ and $\mathcal{M}^{(z)}$, and $D_T$ is used to distinguish $\mathcal{T}$ and $\mathcal{M}^{(z)}$. Using the adversarial loss as the distribution distance measure, we obtain the adversarial losses between $\mathcal{M}^{(z)}$ and $\mathcal{S}$ as
\begin{equation}
\label{eq:adv:st-s}
\begin{split}
\mathcal{L}&_{adv}\left(G_{ST}, D_{S}\right) =\mathbb{E}_{\mathbf{x} \sim P_{S}}\left[\log \left(D_{S}\left(\mathbf{x}\right)\right)\right] \\
&+\mathbb{E}_{\mathbf{x} \sim P_{S}}\left[\log \left(1-D_{S}\left(G_{S T}\left(\mathbf{x}, z\right)\right)\right)\right].
\end{split}
\end{equation}
% \begin{equation}
% \label{eq:adv:st-s}
% \begin{split}
% \mathcal{L}_{adv}\left(G_{ST}, D_{S}\right) =\mathbb{E}_{\mathbf{x} \sim P_{S}}\left[\log \left(D_{S}\left(\mathbf{x}\right)\right)\right] 
% +\mathbb{E}_{\mathbf{x} \sim P_{S}}\left[\log \left(1-D_{S}\left(G_{S T}\left(\mathbf{x}, z\right)\right)\right)\right].
% \end{split}
% \end{equation}

Similarly, the adversarial losses between $\mathcal{M}^{(z)}$ and $\mathcal{T}$ can be written as
\begin{equation}
\label{eq:adv:st-t}
\begin{split}
\mathcal{L}&_{adv} \left(G_{ST}, D_{T}\right) =\mathbb{E}_{\mathbf{y} \sim P_{T}}\left[\log \left(D_{T}\left(\mathbf{y}\right)\right)\right] \\
&+\mathbb{E}_{\mathbf{x} \sim P_{S}}\left[\log \left(1-D_{T}\left(G_{S T}\left(\mathbf{x}, z\right)\right)\right)\right].
\end{split}
\end{equation}
% \begin{equation}
% \label{eq:adv:st-t}
% \begin{split}
% \mathcal{L}_{adv} \left(G_{ST}, D_{T}\right) =\mathbb{E}_{\mathbf{y} \sim P_{T}}\left[\log \left(D_{T}\left(\mathbf{y}\right)\right)\right]
% +\mathbb{E}_{\mathbf{x} \sim P_{S}}\left[\log \left(1-D_{T}\left(G_{S T}\left(\mathbf{x}, z\right)\right)\right)\right].
% \end{split}
% \end{equation}

Deploying the above losses as $\operatorname{dist}\left(P_{S}, P_{M^{(z)}}\right)$ and $\operatorname{dist}\left(P_{T}, P_{M^{(z)}}\right)$ in Eq. (\ref{eq:loss1}), we can derive the following loss
\begin{equation}
\label{eq:adv:st}
\begin{split}
\mathcal{L}_{adv} (G_{ST},& D_S, D_T) = z \cdot \mathcal{L}_{a d v}\left(G_{S T}, D_{T}\right) \\
&+(1-z) \cdot \mathcal{L}_{a d v}\left(G_{S T}, D_{S}\right).
\end{split}
\end{equation}
% \begin{equation}
% \label{eq:adv:st}
% \begin{split}
% \mathcal{L}_{adv} (G_{ST}, D_S, D_T) = z \cdot \mathcal{L}_{a d v}\left(G_{S T}, D_{T}\right)
% +(1-z) \cdot \mathcal{L}_{a d v}\left(G_{S T}, D_{S}\right).
% \end{split}
% \end{equation}

Consider the other direction from $\mathcal{T}$ to $\mathcal{M}^{(1-z)}$, we can define similar loss $\mathcal{L}_{adv} (G_{TS}, D_S, D_T)$.
Then the total adversarial loss is
\begin{equation}
\begin{split}
\mathcal{L}_{adv} &= \mathcal{L}_{adv} (G_{ST}, D_S, D_T) \\
&+ \mathcal{L}_{adv} (G_{TS}, D_S, D_T).
\label{eq:adv_loss}
\end{split}  
\end{equation}
% \begin{equation}
% \begin{split}
% \mathcal{L}_{adv} = \mathcal{L}_{adv} (G_{ST}, D_S, D_T)
% + \mathcal{L}_{adv} (G_{TS}, D_S, D_T).
% \label{eq:adv_loss}
% \end{split}  
% \end{equation}
% and the adversarial training objective is
% \begin{equation}
%     \min_{G_{ST},G_{TS}} \max_{D_S, D_T} \mathcal{L}_{adv}.
% \end{equation}

\paragraph{Modification of Adversarial Loss}
In the loss discussed above, $D_S$ is trained to assign a high value (i.e. 1) to  \textbf{x} and assign a low value (i.e. 0) to $G_{ST}(\mathbf{x}, z)$, and similar for $D_T$. But when $z$ is small, $G_{ST}(\mathbf{x}, z)$ is close to the data from source domain and it will be too aggressive if we train the discriminator to assign 0 to it. In our model, we train the discriminator $D_S$ to assign $1-z$ instead of 0 to $G_{ST}(\mathbf{x}, z)$. When $z=0$, $G_{ST}(\mathbf{x}, z) = \mathbf{x}$ and the discriminator $D_S$ is trained to assign 1 to it. 
$G_{ST}$ and $G_{TS}$ are trained to fool the discriminator $D_S$, trying to make $D_S(G_{TS}(\mathbf{y},z))$ close to 1 and $D_S(G_{ST}(\mathbf{x},z))$ close to $z$.

Besides the adversarial loss, the cycle consistency loss in the cycle GAN here is defined as:
\begin{equation}
\begin{split}
\mathcal{L}&_{\mathrm{cyc}}(G_{ST}, G_{ST}) \\
&=\mathbb{E}_{x \sim P_S}\Vert G_{TS}(G_{ST}(\mathbf{x},z),z)-\mathbf{x}\Vert ^2 \\
&+\mathbb{E}_{\mathbf{y} \sim P_T}\Vert G_{ST}(G_{TS}(\mathbf{y},z),z)-\mathbf{y}\Vert^2.
\end{split}
\label{eq:cyc}
\end{equation}
% \begin{equation}
% \begin{split}
% \mathcal{L}_{\mathrm{cyc}}(G_{ST}, G_{ST}) 
% =\mathbb{E}_{x \sim P_S}\Vert G_{TS}(G_{ST}(\mathbf{x},z),z)-\mathbf{x}\Vert ^2
% +\mathbb{E}_{\mathbf{y} \sim P_T}\Vert G_{ST}(G_{TS}(\mathbf{y},z),z)-\mathbf{y}\Vert^2.
% \end{split}
% \label{eq:cyc}
% \end{equation}

Based on the model structure from \citet{mohiuddin2019revisiting}, we deploy the domain flow on the latent space obtained from two autoencoders, i.e., replace $\mathbf{x}$ and $\mathbf{y}$ with $\text{Enc}_S(\mathbf{x})$ and $\text{Enc}_T(\mathbf{y})$ in above losses. An additional reconstruction loss in the autoencoders is defined as:
% Our model is implemented based on the model structure from \citet{mohiuddin2019revisiting}, in which they apply cycle GAN on the latent features extracted from linear autoencoders. 
% Let $\mathbf{z}_x = \text{Enc}_S(\mathbf{x})$ and $\mathbf{z}_y = \text{Enc}_T(\mathbf{y})$, the adversarial loss and cycle consistency loss that we have discussed before should be deployed on the latent space $Z_x$ and $Z_y$. Besides, we consider an additional reconstruction loss:
\begin{equation}
\label{eq:rec1}
\begin{split}
    L_{rec} &= \mathbb{E}_{x \sim P_S}\Vert \text{Dec}_{S}(\text{Enc}_{S}(\mathbf{x}))-\mathbf{x}\Vert ^2 \\ &+\mathbb{E}_{\mathbf{y} \sim P_T}\Vert \text{Dec}_{T}(\text{Enc}_{T}(\mathbf{y}))-\mathbf{y}\Vert^2.
\end{split}
\end{equation}
% \begin{equation}
% \label{eq:rec1}
% \begin{split}
%     L_{rec} = \mathbb{E}_{x \sim P_S}\Vert \text{Dec}_{S}(\text{Enc}_{S}(\mathbf{x}))-\mathbf{x}\Vert ^2 +\mathbb{E}_{\mathbf{y} \sim P_T}\Vert \text{Dec}_{T}(\text{Enc}_{T}(\mathbf{y}))-\mathbf{y}\Vert^2.
% \end{split}
% \end{equation}

% In practice, Eq. (\ref{eq:rec1}) is used to pretrain the autoencoders before the adversarial training. During the adversarial training, we consider the following reconstruction loss to update autoencoders and generators simultaneously:
% \begin{equation}
% \begin{split}
%     L_{rec} &= \mathbb{E}_{x \sim P_S}\Vert \text{Dec}_{S}(G_{TS}(G_{ST}(\text{Enc}_{S}(\mathbf{x}),z),z))-\mathbf{x}\Vert ^2 \\
%     &+\mathbb{E}_{\mathbf{y} \sim P_T}\Vert \text{Dec}_{T}(G_{ST}(G_{TS}(\text{Enc}_{T}(\mathbf{y}),z),z))-\mathbf{y}\Vert^2.
% \end{split}
% \label{eq:rec2}
% \end{equation}

% The total loss we used is a weighted sum of the three losses we discussed before:
Then the total loss is
\begin{equation}
    L = L_{adv} + \lambda_1 \cdot L_{cyc} + \lambda_2 \cdot L_{rec},
\end{equation}
where $\lambda_1$ and $\lambda_2$ are two hyperparameters.

\paragraph{Choice of $z$}
In our model, $z$ is sampled from a beta distribution $f(z, \alpha, \beta)=\frac{1}{B(\alpha, \beta)} z^{\alpha-1}(1-z)^{\beta-1}$, where $B(\cdot,\cdot)$ is the Beta function, $\beta$ is fixed to be 1, and $\alpha$ is set as a function of the training iterations. Specifically, $\alpha=e^{\frac{t-0.5 T}{0.25 T}}$, where $t$ is the current iteration and $T$ is the total number of iterations. In this setting, $z$ tends to be more likely to be small values at the beginning, and gradually shift to larger values during training. In practice, we set $z=1$ in the last several epochs to fine-tune the model.
For the case of running 10 epochs using our proposed model, the intermediate domain variable z is fixed to be 1 in the last 3 epochs, trying to fine-tune our proposed model. For other cases, i.e., when running 20 and 30 epochs, z is fixed to be 1 in the last 5 epochs.

% embeddings

% drawback of MUSE dictionary:

\section{Bilingual Lexicon Induction}

\subsection{Experimental Setup}

% We use two tasks to evaluate the induced CLWEs. One is the standard intrinsic task, Bilingual Lexical Induction (BLI), and the other is a downstream task, Cross-Lingual Natural Language Inference (XNLI). 

\begin{table*}[htbp]
% \small
\centering
% \left
% \resizebox{0.6\textwidth}{!}{
% \vspace{-11cm}
% \vspace{-0.9cm}
\setlength{\abovecaptionskip}{0.1cm}
\setlength{\belowcaptionskip}{-0.5cm}
\begin{tabular}{lllll}\hline\toprule
Languages & Language Family & Morphology Type & Wikipedia Size & ISO 639-1 \\
\hline
English & Indo-European & isolating & 6.46M & en \\
\hline
German & Indo-European & fusional & 2.67M & de \\
French & Indo-European & fusional & 2.40M & fr \\
Russian  & Indo-European & fusional & 1.80M & ru  \\
Spanish  & Indo-European & fusional & 1.75M & es \\
Italian & Indo-European & fusional & 1.74M & it \\
Japanese & Japonic & agglutinative & 1.31M & ja \\
Chinese & Sino-Tibetan & isolating & 1.25M & zh \\
Arabic & Afro-Asiatic & fusional & 1.16M & ar \\
Finnish & Uralic & agglutinative & 0.52M & fi \\
Turkish & Turkic & agglutinative & 0.47M & tr \\
Malay & Austronesian & agglutinative & 0.36M & ms\\
Hebrew & Afro-Asiatic & fusional & 0.31M & he \\
Bulgarian & Indo-European & isolating & 0.28M & bg \\
Hindi & Indo-European & fusional & 0.15M & hi \\
\hline\toprule
\end{tabular}
% }
\caption{The list of 15 languages from our main BLI experiments along with their corresponding language family, broad morphological type, number of wikipedia articles 
% \tablefootnote{https://en.wikipedia.org/wiki/List\_of\_Wikipedias\#Details\_table}
of monolingual corpora, and their ISO 639-1 codes
% \tablefootnote{https://en.wikipedia.org/wiki/List\_of\_ISO\_639-1\_codes}
.}
\label{table:lang}
\end{table*}

Bilingual Lexical Induction (BLI) has become the \textit{de facto} standard evaluation for mapping-based CLWEs (\citealp{ruder2019survey}; \citealp{glavas2019properly}; \citealp{vulic2019we}).
Given a shared CLWE space and a list of source language words, the task is to retrieve their target translations based on their word vectors.
The lightweight nature of BLI allows us to conduct a comprehensive evaluation across a large number of language pairs. 
We adopt Cross-domain Similarity Local Scaling (CSLS) from \citet{conneau2017word} for the nearest neighbor retrieval.
Following a standard evaluation practice (\citealp{mikolov2013exploiting}; \citealp{conneau2017word}; \citealp{mohiuddin2019revisiting}), we evaluate the performance of induced CLWEs with the Precision at k (P@k) metric, which measures the percentage of ground-truth translations that are among the top k ranked candidates.

\paragraph{Monolingual Embeddings} 
Following prior works (\citealp{conneau2017word}; \citealp{sogaard2018limitations}; \citealp{glavas2019properly}; \citealp{mohiuddin2019revisiting}), we use 300-dimensional fastText embeddings \citep{bojanowski2017enriching} \footnote{https://fasttext.cc/docs/en/pretrained-vectors.html} trained on Wikipedia monolingual corpus. All vocabularies are truncated to the 200K most frequent words.

\paragraph{Evaluation Dictionary} 
For the BLI task, we evaluate our model on two datasets. The first one is from \citet{conneau2017word}, which contains 110 bilingual dictionaries.\footnote{https://github.com/facebookresearch/MUSE\#ground-truth-bilingual-dictionaries}
The other one is from \citet{dinu2014improving} and its extension by \citet{artetxe2018generalizing}, which consists of gold dictionaries for 4 language pairs. This dataset is refered to as \textbf{Dinu-Artetxe} in later sections.

\paragraph{Selection of Languages} 

From the 110 bilingual test dictionaries of \citet{conneau2017word}, we select test language pairs based on the following goals: a) for a direct comparison with recent works, we aim to cover both close and distant language pairs that are discussed in recent works; and b) for analyzing a large set of language pairs, we aim to ensure the coverage of different language properties and training data size. The resulting 15 languages (including English) are listed in Table \ref{table:lang}\footnote{Since the training corpus size of fastText embeddings is not available, the Wikipedia Size in Table \ref{table:lang} is chosen as the number of wikipedia articles listed in https://en.wikipedia.org/wiki/List\_of\_Wikipedias\#Details\_table. It may differ from the actual size of fastText training corpus but the relative size among languages should not be far away from this.}. We run BLI evaluations for language pairs of English and the rest 14 languages in both directions. The first five languages other than English in Table \ref{table:lang} belong to the same language family as English and have large monolingual corpora, which are categorized as "close" languages to English. The remaining languages are then categorized as "distant" to English since they belong to different language family from English or have much smaller monolingual corpora.
For the \textbf{Dinu-Artetxe} dataset, we select all its 4 language pairs, i.e., English-German (en-de), English-Spanish (en-es), English-Italian (en-it), and English-Finnish (en-fi).

% how to define "distant": 
% - language family + morphological type (\citet{sogaard2018limitations}; \citet{vulic2019we})

% how to define "low-resource":
% - wikipedia size (\cite{vulic2020all}; \citet{doval2019robustness})

\begin{table*}[htbp]
% \small
\centering
% \resizebox{0.6\textwidth}{!}{
% \vspace{-11cm}
\vspace{-0.5cm}
\setlength{\abovecaptionskip}{0cm}
\setlength{\belowcaptionskip}{-0.5cm}
\begin{tabular}{c|cc|cc|cc|cc|cc}\hline\toprule
models & en-de & de-en & en-fr & fr-en & en-ru & ru-en & en-es & es-en & en-it & it-en \\
\hline
\text {MUSE } & 75.26 & 72.53 & 82.21 & 82.36 & 40.65 & 58.46 & \textbf{82.60} & 83.40 & 77.87 & 77.30 \\
\text {VecMap} & 75.33 & 74.00 & 82.20 & \textbf{83.73} & 47.13 & 64.13 & 82.33 & 84.33  & 78.67 & 79.40   \\
\text{\textbf{AutoEnc}}  & 75.77 & 73.87 & \textbf{82.55} & 83.33 &47.29 & 64.48 & 82.25 & 84.21 & 79.06 & \textbf{80.20} \\
\hline
\text{Our model} & \textbf{75.78} & \textbf{74.83} & 82.53 & 83.51 & \textbf{47.55} & \textbf{65.08} & 82.36 & \textbf{84.43} & \textbf{79.10} & \textbf{80.20} \\
\hline\toprule
\end{tabular}
% }
\caption{BLI translation precision (P@1) of \citet{conneau2017word} on close language pairs. The highest precision values are boldfaced.}\label{res:prec_sim}
\end{table*}

\paragraph{Baselines}
We compare our model with the well-known unsupervised models of \citet{conneau2017word} (MUSE)\footnote{https://github.com/facebookresearch/MUSE} and \citet{artetxe2018robust} (VecMap)\footnote{https://github.com/artetxem/vecmap}.
% We also compare our model with \citet{mohiuddin2019revisiting}\footnote{https://ntunlpsg.github.io/project/unsup-word-translation/} as the reference baseline. For simplicity, we use \textbf{AutoEnc} to denote the reference model.
We implemented our proposed method of domain flow based on \citet{mohiuddin2019revisiting}\footnote{https://ntunlpsg.github.io/project/unsup-word-translation/}, which is denoted as \textbf{AutoEnc} in this paper.
% We also compare our model with \citet{mohiuddin2019revisiting}\footnote{https://ntunlpsg.github.io/project/unsup-word-translation/}, based on which we implement the domain flow. For simplicity, we use \textbf{AutoEnc} to denote it.
Results of all baselines are obtained by rerunning the public codes with the default settings on our machine. 
More experimental details are presented in Appendix \ref{exp_detail}.\footnote{We have not compared with \citet{li2020simple} since its code is not publicly available.}

% \section{Results}

% \subsection{BLI Results}
\subsection{Results and Discussion}

% \subsubsection{Results on Precision}
% \subsubsection{Results on Precision}
\paragraph{Results of Precision on \citet{conneau2017word} Dataset}

We report the average precision at 1  (across the successful runs out of 10 runs) on \citet{conneau2017word} dataset in Table \ref{res:prec_sim}, Table \ref{res:prec_dis1}, and Table \ref{res:prec_dis2}. The P@1 of close language pairs is shown in Table \ref{res:prec_sim} and that of distant language pairs is Table \ref{res:prec_dis1} and Table \ref{res:prec_dis2}.
% The best precision of \textbf{AutoEnc} and the proposed model is also added in Table \ref{res:prec_dis}. 
The detailed results including P@5, P@10 are presented in Appendix \ref{appendix:bli}.
As shown in Table  \ref{res:prec_sim}, the proposed model performs on par with that of the unsupervised baselines in close language pairs.
As shown in Table \ref{res:prec_dis1} and Table \ref{res:prec_dis2}, the proposed model gets comparable results with the baselines on distant language pairs en-bg and en-ms and achieves higher precision on en-he, en-tr, en-hi, en-fi, and en-ar.  For more distant language pairs en-ja and en-zh, AutoEnc fails for all 10 runs.\footnote{Kindly note that we obtained one successful run for en-zh and 2 for en-ja when we changed the default SGD optimizer of the released code to Adam, with the learning rate of 0.001, and trained 50 epochs with epoch size of 100,000 for 20 times.}
% Note that we got 1 successful run for en-zh and 2 for en-ja if we changed its default SGD optimizer to adam with the learning rate of 0.001 and trained 50 epochs with epoch size of 100,000 for 20 times.
Even for the most robust system VecMap, it still fails on en-zh. Our results show that the proposed model offers enhanced robustness that is able to complete these two difficult cases and exceed the performance of all baselines.

\begin{table*}[htbp]
% \small
\normalsize
\centering
% \resizebox{0.9\textwidth}{!}{
% \vspace{-11cm}
% \vspace{-0.5cm}
\setlength{\abovecaptionskip}{0.1cm}
\setlength{\belowcaptionskip}{-0.1cm}
\begin{tabular}{c|cc|cc|cc|cc|cc}\hline\toprule
models & en-he & he-en  & en-tr & tr-en & en-bg & bg-en & en-hi & hi-en & en-fi & fi-en \\
\hline
\text {MUSE } & 36.26 & 54.07 & 45.40 & 59.65 & 41.97 & 56.29 & 30.40 & 39.55  & 42.70 & 59.30 \\
\text {VecMap}  & 41.20 & 52.80 & 48.06 & 59.64 & 44.46 & 57.80 & 35.33 & 37.57  & 46.13 & 62.27\\
\text{\textbf{AutoEnc}} & 43.71 & 57.76  & 49.31 & 61.01 & 46.62 & \textbf{61.01} & 35.74 & 46.58 & 44.65 & 62.59 \\
\hline
\text{Our model} & \textbf{44.67} & \textbf{58.01} & \textbf{50.45} & \textbf{61.19} & \textbf{46.77} & 60.85 & \textbf{36.88} & \textbf{49.17}  & \textbf{48.28} & \textbf{64.66} \\
% \hline
% \text{\textbf{AutoEnc} (best)} & 45.13 & 58.21 & 38.20 & 55.35 & - & -  & 0 & 0 & 0 &0\\
% \text{Our model (best)} & \textbf{46.27} & \textbf{59.14} & \textbf{38.93} & \textbf{56.09} & -  & - & \textbf{48.93} & \textbf{33.91} & \textbf{39.47} & \textbf{34.07}\\
\hline\toprule
\end{tabular}
% }
\caption{BLI translation precision (P@1) of \citet{conneau2017word} on distant language pairs. The highest precision values are boldfaced.}\label{res:prec_dis1}
\end{table*}

\begin{table*}[htbp]
% \small
\normalsize
\centering
% \resizebox{0.9\textwidth}{!}{
% \vspace{-11cm}
% \vspace{-0.5cm}
\setlength{\abovecaptionskip}{0.1cm}
\setlength{\belowcaptionskip}{-0.4cm}
\begin{tabular}{c|cccc|cccc}\hline\toprule
models & en-ar & ar-en & en-ms & ms-en  & en-ja & ja-en & en-zh & zh-en\\
\hline
\text {MUSE }  & 32.00 & 0  & 0.67 & 0.67  & 0 & 5.31 & 0.67 & 0\\
\text {VecMap}  & 34.79 & 49.26 & 48.40 & 38.80 & 37.56 & 26.12 & 0 & 0\\
\text{\textbf{AutoEnc}} & 36.27 & 53.27 & \textbf{53.42} & 48.53  & 0 & 0 & 0 &0\\
\hline
\text{Our model}  & \textbf{36.33} & \textbf{53.60}   & 52.64 & \textbf{51.22} & \textbf{46.67} & \textbf{31.11} & \textbf{41.45} & \textbf{32.62}\\
% \hline
% \text{\textbf{AutoEnc} (best)} & 38.20 & 55.35 & \textbf{55.20} & 52.30  & 0 & 0 & 0 &0\\
% \text{Our model (best)} & \textbf{38.93} & \textbf{56.09} & 54.60  & \textbf{53.24} & \textbf{48.93} & \textbf{33.91} & \textbf{39.47} & \textbf{34.07}\\
\hline\toprule
\end{tabular}
% }
\caption{BLI translation precision (P@1) of \citet{conneau2017word} on distant language pairs. The highest precision values are boldfaced.}\label{res:prec_dis2}
\end{table*}

\paragraph{Results of Precision on \textbf{Dinu-Artetxe} Dataset}
We report the average precision at 1 on \textbf{Dinu-Artetxe} dataset in Table \ref{res:prec_dinu}. Similar conclusion can be drawn. Specifically, the proposed model gets comparable results with baselines for close language pairs en-de, en-es, and en-it. Moreover, the proposed model achieves higher precision on distant-language pair en-fi.

\begin{table}[!htbp]
% \small
\normalsize
\centering
% \resizebox{0.9\textwidth}{!}{
% \vspace{-11cm}
% \vspace{-0.2cm}
\setlength{\abovecaptionskip}{0.1cm}
\setlength{\belowcaptionskip}{-0.5cm}
\begin{tabular}{c|cccc}\hline\toprule
models & en-de & en-es & en-it & en-fi\\
\hline
\text {MUSE }  & 65.57 & 67.89  & 67.45 & 30.76 \\
\text {VecMap}  & 65.73 & 69.31 & 68.71 & 37.95\\
\text{\textbf{AutoEnc}} & 65.76 & 69.25 & 68.76 & 36.38 \\
\hline
\text{Our model}  & \textbf{66.01} & \textbf{69.53}  & \textbf{68.78} & \textbf{38.35} \\
\hline\toprule
\end{tabular}
% }
\caption{BLI translation precision (P@1) of \textbf{Dinu-Artetxe} dataset on all 4 language pairs. The highest precision values are boldfaced.}
\label{res:prec_dinu}
\end{table}

% More results on precision are shown in Appendix \ref{res_pre}.

\paragraph{Results on Robustness}
% \subsubsection{Results on Robustness}
% \subsection{Results on Robustness}

We reference \citet{artetxe2018robust} to define that a system ‘succeeds’ when it attains a precision above 5\% and ‘fails’ otherwise. The precision in above sections is averaged over all successful runs. Indeed, the baselines do not always succeed even in the case where a positive precision is reported. For instance, using the default setting (5 epochs with epoch size of 1,000,000) and running the codes 10 times, \textbf{AutoEnc} model succeeds 7 runs on en-he, 7 runs on en-ar, and 5 runs on en-hi. Even worse, out of 10 runs, the baseline MUSE only succeeds 6 runs on en-he, 7 runs on en-ar, and 2 runs on en-hi. To show the effectiveness of the proposed domain flow in improving the robustness of adversarial models, we compare the proposed model with \textbf{AutoEnc} on distant language pairs of en-he, en-tr, en-bg, en-hi, en-fi, en-ar and en-ms.\footnote{En-ja and en-zh are not compared since Table \ref{res:prec_dis2} has shown that \textbf{AutoEnc} can never succeed in these two language pairs.} Specifically, using an epoch size of 100,000 and various numbers of epochs, we count the successful runs of the proposed model and \textbf{AutoEnc} out of 10 runs. The results are shown in Table \ref{res:rob}. As shown in Table \ref{res:rob}, our model succeeds with a much higher probability (cf \textbf{AutoEnc}) for all the 3 language pairs, as well as succeeds 100\% if it is given 30 epochs. 
Note that the default setting of \textbf{AutoEnc} is equivalent to 50 epochs. 
% Note that the default setting of \textbf{AutoEnc} is 5 epochs with epoch size of 1,000,000, which is equivalent to 50 epochs in our setting. 
\textbf{AutoEnc} can not achieve 100\% success under this setting. However, for the proposed approach, 30 epochs are sufficient for all the 7 distant language pairs.

\begin{table*}[!h]
\centering
\vspace{-0.25cm}
\setlength{\abovecaptionskip}{0.2cm}
\setlength{\belowcaptionskip}{-0.4cm}
% \resizebox{0.45\textwidth}{!}{
\begin{tabular}{c|c|ccccccc}
\hline
\hline
$\#$ of epochs & models & en-he & en-tr & en-bg & en-hi & en-fi & en-ar & en-ms \\
\hline 
\multirow{2}{*}{10} &\textbf{AutoEnc} & 2 & 7 & 8 & 6 & 2 & 3  & 0\\
& \text{Ours}  & 6 & 9 & 10 & 9 & 6 & 8 & 9\\
\hline 
\multirow{2}{*}{20} &\textbf{AutoEnc}  & 6 & 7 & 8 & 7 & 4 & 6 & 2\\
&\text{Ours} &8 & 10 & 10 & 8 & 10 & 10 &10\\
\hline 
\multirow{2}{*}{30} &\textbf{AutoEnc}  & 7 & 7 & 9 & 7 & 4 & 7 & 4\\
& \text{Ours}  & 10 & 10 & 10 & 10 & 10 & 10 & 10 \\
\hline\hline
\end{tabular}
% }
\caption{Number of successful runs out of 10 runs.}\label{res:rob}
\end{table*}

\paragraph{Visualization on Smoothing Training}
% \subsubsection{Visualization on Smoothing Training}
% \subsection{Visualization on Smoothing Training}

\begin{figure*}[htbp]
% \vspace{-0.3cm}
\setlength{\abovecaptionskip}{0cm} 
\setlength{\belowcaptionskip}{-0.4cm}
\centering
\subfigure[\textbf{AutoEnc} - without domain flow]{
\centering
\includegraphics[width=2.8in]{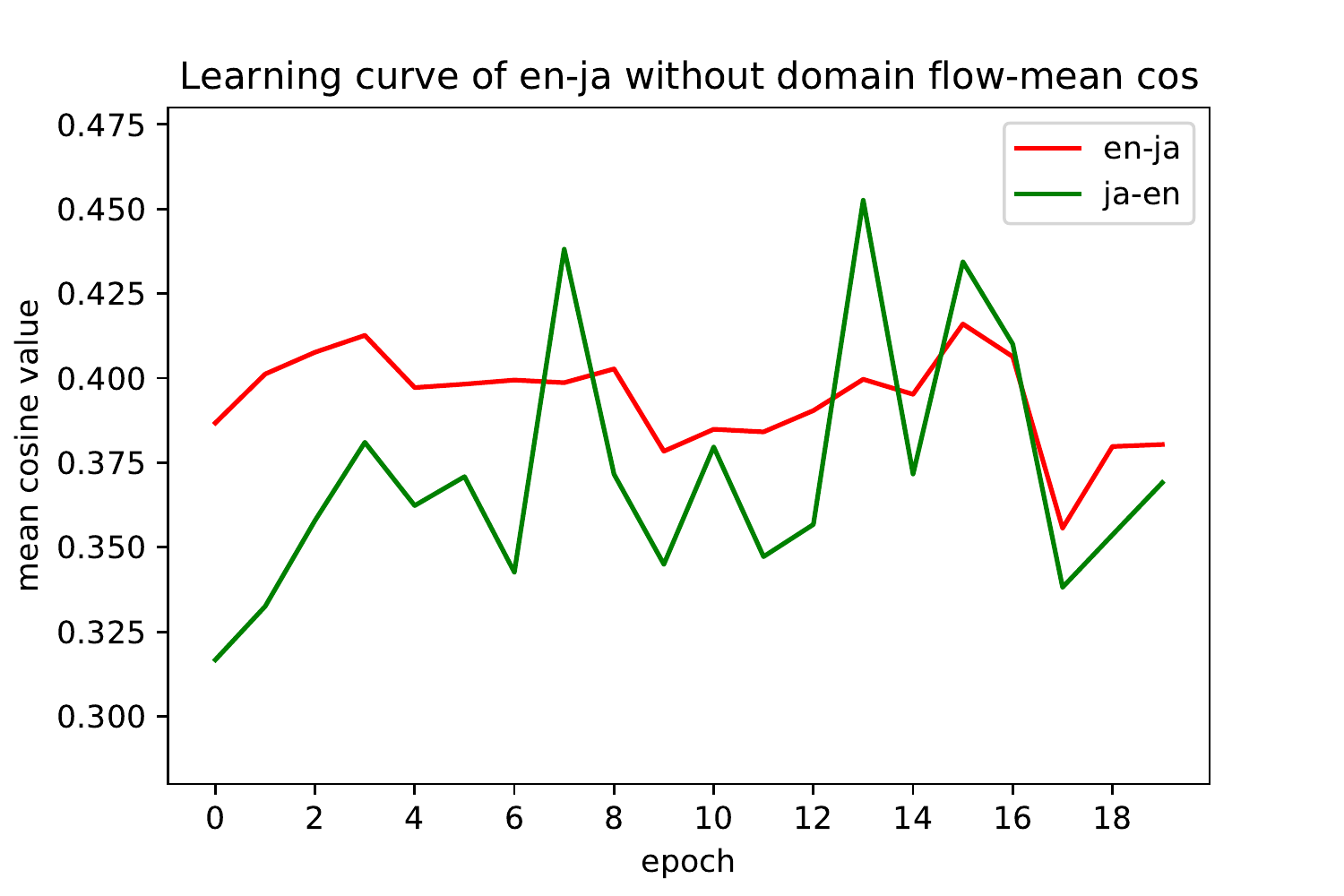}
}%
\quad
\subfigure[Proposed model - with domain flow]{
\centering
\includegraphics[width=2.8in]{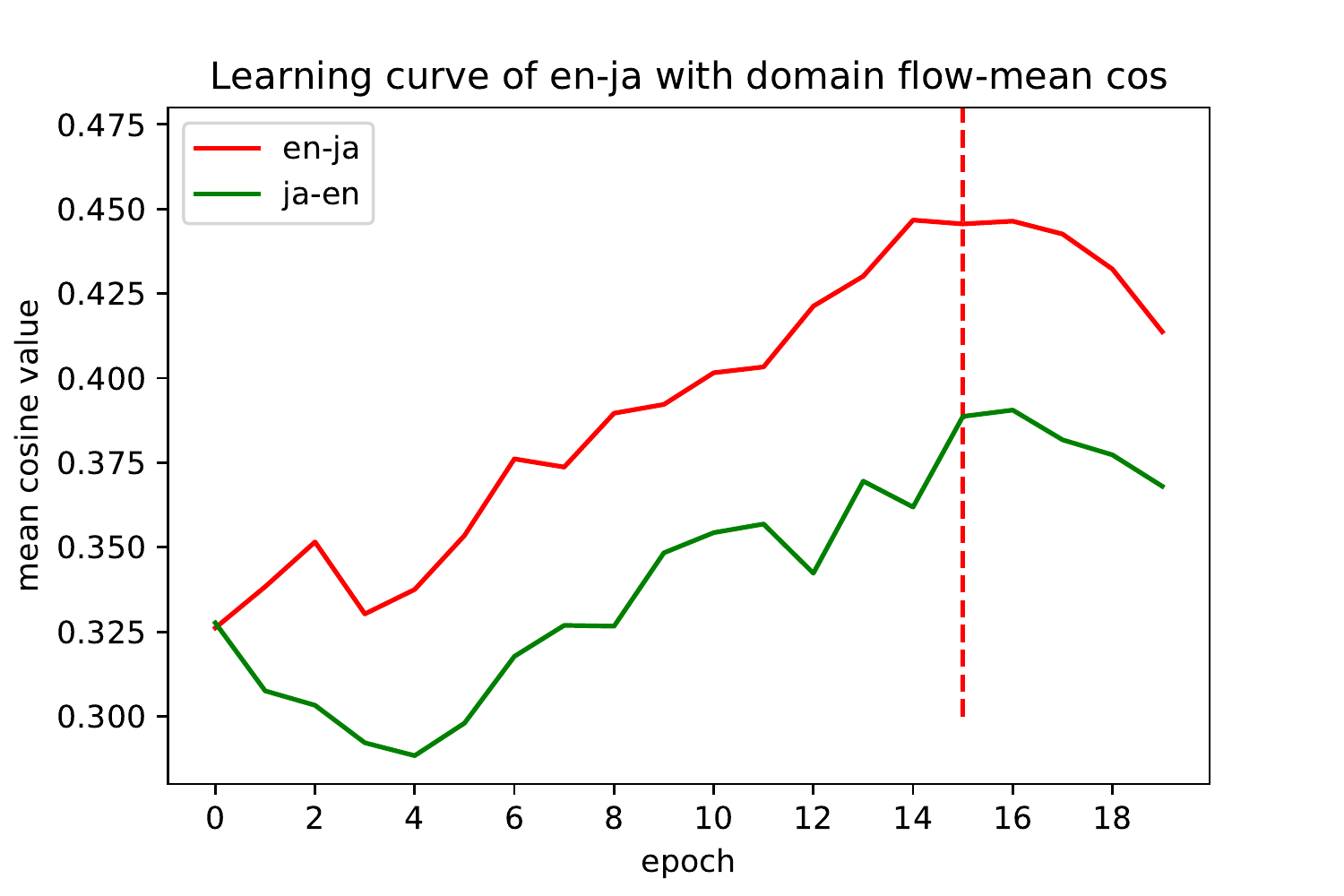}
}%
\quad
\centering
\caption{Learning curves of en-ja: mean cosine value in a pseudo dictionary (Appendix \ref{valid_crit}) w.r.t. training iterations.}
\label{fig:en-ja}
\end{figure*}

We use the unsupervised validation criterion proposed by \citet{conneau2017word}, which is the mean cosine value in a pseudo dictionary (details in Appendix \ref{valid_crit}), to show the learning curves of en-ja in Figure \ref{fig:en-ja}.
% The figure is an illustration with the language pair en-ja, where 
We observe that the training procedure of \textbf{AutoEnc} is very unstable. However, the mean cosine value of the proposed model can increase smoothly (z=1 in the last 5 epochs, i.e., applying  5 training iterations of \textbf{AutoEnc} to do fine-tuning). This suggests that the proposed model can learn cross-lingual mappings along a smoother trajectory than \textbf{AutoEnc}. This demonstrates the advantage of introducing intermediate domains to smoothly bridge the gap between the source and target domains, thus improving the robustness of unsupervised models.

\section{Cross-Lingual Natural Language Inference}

Moving beyond the BLI evaluation, we evaluate our model on a language understanding downstream task, Cross-Lingual Natural Language Inference (XNLI). Given a pair of sentences, the Natural Language Inference (NLI) task is to detect entailment, contradiction and neutral relations between them. We test our model on a zero-shot cross-lingual transfer setting where an NLI model is trained on English corpus and then tested on a second language (L2). 
Following \citet{glavas2019properly}, we train a well-known robust neural NLI model, Enhanced Sequential Inference Model (ESIM; \citealp{chen2016enhanced}) on the English MultiNLI corpus \citep{williams2017broad}.\footnote{https://cims.nyu.edu/\~{}sbowman/multinli/} The English word embeddings are from the induced CLWE space and are kept fixed during training. The trained ESIM model is then evaluated on the L2 portion of the XNLI \citep{conneau2018xnli}\footnote{https://github.com/facebookresearch/XNLI}  by changing the embedding layer to L2 embeddings from the induced CLWE space. The test languages are the intersection of the selected 14 languages in Table \ref{table:lang} and the available languages in XNLI dataset, which include 9 languages, i.e., German(de), French(fr), Russian(ru), Spanish(es), Chinese(zh), Arabic(ar), Turkish(tr), Bulgarian(bg), and Hindi (hi).

% drawback of MUSE:
% words from the training dictionary
% have higher frequencies than words from the test set\footnote{https://github.com/facebookresearch/MUSE/issues/24}
% [Why Overfitting Isn’t Always Bad: Retrofitting Cross-Lingual Word Embeddings to Dictionaries]

% \subsection{XNLI Results}
\subsection{Results and Discussion}

We report the average XNLI accuracy scores (from 3 NLI model training) in Table \ref{table:xnli} and Table \ref{table:xnli2}. The accuracy values in Table \ref{table:xnli} are obtained by evaluating the L2 portion of XNLI using the fully trained English NLI model. We also evaluate the XNLI accuracy of L2 using the English NLI model checkpoint from every training epoch. We observe that while the test accuracy of English on MNLI is always increasing with more training epochs, the XNLI accuracy of L2 increases in the first few epochs and then decreases. The phenomenon may come from the overfitting of English NLI model. Thus we also report the highest XNLI accuracy value of L2 in Table \ref{table:xnli2} by evaluating the L2 portion of XNLI on every checkpoint and selecting the highest one. Following \citet{glavas2019properly}, we use asterisks to denote language pairs for which the given CLWE models sometimes could not yield successful runs in BLI task.

\begin{table*}[htbp]
% \small
\normalsize
\centering
% \resizebox{0.9\textwidth}{!}{
% \vspace{-11cm}
% \vspace{-0.7cm}
\setlength{\abovecaptionskip}{0.2cm}
\setlength{\belowcaptionskip}{-0.1cm}
\begin{tabular}{c|c|c|c|c|c|c|c|c|cc}\hline\toprule
models & de & ru & es & fr & ar & tr & bg & hi & zh \\
\hline
\text {MUSE } & \textbf{56.53} & 47.33 & 48.75 & 41.40 & 45.36* & 50.18 & 47.10* & 34.04*  & 34.06* \\
\text {VecMap}  & 51.38 & 45.46 & 40.51 & 36.61 & 41.13 & 44.74 & 41.60 & 37.43  & 33.89* \\
\text{\textbf{AutoEnc}} & 56.21 & \textbf{49.73} & \textbf{50.45}  & 40.83 & 43.80* & 51.56 & 48.83* & 34.91* & 33.67* \\
\hline
\text{Our model} & 56.33 & 49.00 & 49.32 & \textbf{42.50} & \textbf{51.64} & \textbf{52.28} & \textbf{51.82} & \textbf{43.47}  & \textbf{45.44} \\
% \hline
% \text{\textbf{AutoEnc} (best)} & 45.13 & 58.21 & 38.20 & 55.35 & - & -  & 0 & 0 & 0 &0\\
% \text{Our model (best)} & \textbf{46.27} & \textbf{59.14} & \textbf{38.93} & \textbf{56.09} & -  & - & \textbf{48.93} & \textbf{33.91} & \textbf{39.47} & \textbf{34.07}\\
\hline\toprule
\end{tabular}
% }
\caption{XNLI performance (test set accuracy) from fully trained English NLI model. The highest accuracy values are boldfaced. Asterisks denote language pairs for which CLWE models sometimes could not yield successful runs in the BLI tasks.}\label{table:xnli}
\end{table*}

\begin{table*}[htbp]
% \small
\normalsize
\centering
% \resizebox{0.9\textwidth}{!}{
% \vspace{-11cm}
% \vspace{-0.7cm}
\setlength{\abovecaptionskip}{0.2cm}
\setlength{\belowcaptionskip}{-0.3cm}
\begin{tabular}{c|c|c|c|c|c|c|c|c|cc}\hline\toprule
models & de & ru & es & fr & ar & tr & bg & hi & zh \\
\hline
\text {MUSE } & 58.65 & 48.81 & 49.87 & 42.82 & 47.07* & 51.78 & 49.27* & 34.65*  & 34.89* \\
\text {VecMap}  & 52.46 & 46.35 & 41.45 & 38.13 & 43.54 & 45.50 & 44.58 & 43.83  & 34.61* \\
\text{\textbf{AutoEnc}} & \textbf{60.23}  & 51.36 & \textbf{51.63} & 42.94 & 46.27* & 53.92 & 50.46* & 39.47* & 34.67* \\
\hline
\text{Our model} & 59.73  & \textbf{51.50} & 51.36 & \textbf{46.40} & \textbf{55.13} & \textbf{54.18} & \textbf{54.92} & \textbf{46.11}  & \textbf{50.76} \\
% \hline
% \text{\textbf{AutoEnc} (best)} & 45.13 & 58.21 & 38.20 & 55.35 & - & -  & 0 & 0 & 0 &0\\
% \text{Our model (best)} & \textbf{46.27} & \textbf{59.14} & \textbf{38.93} & \textbf{56.09} & -  & - & \textbf{48.93} & \textbf{33.91} & \textbf{39.47} & \textbf{34.07}\\
\hline\toprule
\end{tabular}
% }
\caption{Highest XNLI performance (test set accuracy) during the English NLI model training. The highest accuracy values are boldfaced. Asterisks denote language pairs for which CLWE models sometimes could not yield successful runs in the BLI task.}\label{table:xnli2}
\end{table*}

As shown in Table \ref{table:xnli} and Table \ref{table:xnli2}, the proposed model gets comparable results on close languages de, ru, and es. For all other languages, the proposed model outperforms all the baselines. Moreover, the performance of the proposed model gives marked improvements over all baselines for the languages where CLWE models yield failed runs in BLI task. When testing in zh, the XNLI accuracy values of all baselines are close to 33.33\%, which means their predictions are almost like random guessing. Furthermore, the test XNLI accuracy values for NLI models trained from "failed" CLWEs are all close to 33.33\% , which degrades the overall accuracy of ar, bg, and hi (shown in Table \ref{table:xnli} and Table \ref{table:xnli2}). This observation validates the importance of successful runs in BLI for downstream XNLI task and the high accuracy of the proposed model demonstrate its effectiveness in improving the robustness of adversarial models.

\section{Related Work}

% check \cite{bhowmik2021leveraging}

\paragraph{Adversarial Approaches}
Adversarial training has shown great success in inducing unsupervised CLWEs. This was first proposed by \citet{barone2016towards}, who initially utilized an adversarial autoencoder to learn CLWEs. Despite encouraging results, his model is not competitive with approaches using bilingual seeds. 
Based on a similar model structure, \citet{zhang2017adversarial} improve the adversarial training with orthogonal parameterization and cycle consistency. They incorporate additional techniques like noise injection to aid training and report competitive results on the BLI task.
Their follow-up work \citep{zhang2017earth} propose to view word embedding spaces as distributions and to minimize their earth mover's distance. 
\citet{conneau2017word} report impressive results on a large BLI dataset by adversarial training and an iterative refinement process. 
\citet{mohiuddin2019revisiting} revisit the adversarial training and focus on mitigating the robustness issue by adding cycle consistency loss and learning cross-lingual mapping in a latent space encoder by autoencoders.
% \citet{alvarez2018gromov}

\paragraph{Projecting Embeddings into Intermediate Spaces}
The idea of projecting word embedding spaces into intermediate spaces has been explored in other works. 
\citet{jawanpuria2019learning} propose to decouple the cross-lingual mappings into language-specific rotations, to align embeddings in a latent common space, and a language-independent similarity metric, i.e., Mahalanobis metric, to measure the similarity of words in the latent common space. 
\citet{doval2018improving} apply an additional transformation to refine the already aligned word embeddings, which moves cross-lingual synonyms towards a middle point between them. 
% The authors empirically demonstrated that there are significant gaps in the cross-lingual space between words and their translated equivalents. Their method directly attempts to reduce these gaps by moving each word vector towards the middle point between its current representation and the representation of its translation.
\citet{kementchedjhieva2018generalizing} propose to project two languages onto a third latent space via generalized procrustes analysis. Different from us, their work focus on weakly supervised settings and they consider no constraint on the third latent space. However, we focus on the unsupervised case and the third space in our model is forced to approach the target space gradually staring from the source space.
\citet{heyman2019learning} utilize a similar idea to induce multilingual word embeddings. To learn a shared multilingual embedding space for a variable number of languages, they propose to incrementally add new languages to the current multilingual space. They find that it is beneficial to project close languages first and then more distant languages. Different from our work, their model is based on real languages and requires to construct a language order for the incremental learning. In contrast, our model utilizes a sequence of pseudo-languages and no additional engineering work on language order is required.

\paragraph{Motivation for unsupervised CLWE}
Despite the success of fully unsupervised CLWE approaches, some recent works have questioned the motivations behind them. \citet{sogaard2018limitations} challenge the basic assumption of unsupervised CLWE approaches that monolingual word embedding graphs are approximately isomorphic. They further show the well-known adversarial approach, MUSE \citep{conneau2017word}, performs poorly on morphologically rich languages and adding a weak supervision signal from identical words enables more robust induction. \citet{vulic2019we} show that the most robust non-adversarial approach, VecMap \citep{artetxe2018robust}, still fails on 87 out of 210 distant language pairs. \citet{artetxe2020call} argue that a scenario without any parallel data and abundant monolingual data is unrealistic in practice. However, \citet{sogaard2018limitations} shows that unsupervised approaches can sometimes outperform the supervised approaches (last row in their Table 2). \citet{vulic2019we} point out that techniques learned from unsupervised approaches can also benefit supervised and weakly supervised approaches. \citet{artetxe2020call} analyze the scientific value of unsupervised approaches. According to the literature, it is still an open problem which approach is better. The main drawback of unsupervised approaches lies in their instability. The research value of unsupervised approaches may exceed that of supervised approaches if we can solve the instability problem. \footnote{Additional discussions on CLWE and multilingual pre-trained language models are in Appendix \ref{clwe_mplm}.}

% \paragraph{Domain Flow} Domain flow has shown effectiveness in domain adaptation tasks in the computer vision field. 
% \citet{cui2014flowing} learn domain-invariant features by representing the domains via interpolating intermediate points on a Riemann manifold by characterizing samples from each domain as a covariance matrix. 
% \citet{gong2019dlow} introduced intermediate domains to generate images of intermediate styles. They added an intermediate domain variable on the input of the generator via conditional instance normalization. The intermediate domains can smoothly bridge the gap between source and target domains to ease the domain adaptation task.
% Inspired by this idea, we propose to adopt domain flow for the unsupervised cross-lingual word embedding task. 

\section{Conclusion}

% In this paper, we provide a new perspective on learning cross-lingual word embedding via domain flow. Experimental results show the effectiveness of our model in improving the robustness of adversarial models and smoothing the training stage. Comparison with state-of-the-art pretrained cross-lingual models show the advantages of our model in constructing a universal embedding space that is spatially structured according to semantics.  

% This paper proposes an approach for improving the robustness in unsupervised learning of cross-lingual word embeddings by leveraging the idea of domain flow from computer vision. Experimental results demonstrate that the proposed approach can effectively enhance the robustness of adversarial models during training, together with a smooth learning curve.

This paper proposes an approach for improving the robustness of unsupervised cross-lingual word embeddings by leveraging the idea of domain flow from computer vision. 
Experimental results on BLI tasks demonstrate that the proposed approach can achieve comparable precision on close language pairs and effectively enhance the robustness of adversarial models on distant language pairs, achieving a smooth learning curve. The experiments on XNLI tasks further validate importance of successful runs in BLI and the effectiveness of the proposed model.
% In future work, we will focus on improving the translation precision of distant language pairs.

% This paper proposes the approach of learning cross-lingual word embedding by leveraging the idea of domain flow from computer vision. Experimental results show that the proposed approach effectively enhances the robustness of adversarial models during training, together with a smooth learning curve. 

% \section*{Acknowledgements}

% Additional elements were taken from the formatting instructions of the \emph{International Joint Conference on Artificial Intelligence} and the \emph{Conference on Computer Vision and Pattern Recognition}.

\bibliography{anthology,custom}
\bibliographystyle{acl_natbib}

\newpage
% ~
% \newpage

\appendix
% \begin{appendices}
\renewcommand{\appendixname}{Appendix ~\Alph{section}}
\section{Experimental Details}
\label{exp_detail}

All experiments are conducted on a single GPU of Tesla V100 DGXS 32GB.

% \paragraph{Datasets}
% The monolingual embeddings are chosen as 300-dimensional fastText embeddings\footnote{https://fasttext.cc/docs/en/pretrained-vectors.html} trained on Wikipedia monolingual corpus by \citet{conneau2017word}.
% The test dictionary is also taken from \citet{conneau2017word}, in which each test set consists of 1500 word translation pairs. 

\paragraph{Parameter and Optimizer Settings}
In our setting, $\lambda_1$ and $\lambda_2$ are set to be 5 and 1, respectively. The dimension of the hidden space from autoencoders is set to be 350. We use the Adam optimizer \citep{kingma2014adam} with the learning rate of 0.001 and the batch size of 32 to train our model. The epoch size is chosen as 100,000. Note that in the default setting of MUSE and \textbf{AutoEnc}, the epoch size is chosen as 1,000,000 and the optimizer is chosen as the SGD with the learning rate of 0.1.

\paragraph{Fine-tuning Steps}
After the initial mapping is derived from the adversarial training, some fine-tuning steps are used to refine it, just like what \citet{conneau2017word} and \citet{mohiuddin2019revisiting} did in their work. That is, using the self-learning scheme to update the mappings and dictionaries alternatively and apply symmetric re-weighting \citep{artetxe2018generalizing} on the transformed word embeddings.
Then Cross-domain Similarity Local Scaling \citep{conneau2017word} is used to retrieve the word translation.

\section{Detailed BLI Results}
\label{appendix:bli}

The detailed BLI precision results on \citet{conneau2017word} dataset are shown in Table \ref{table:prec_all_muse} and the results on \textbf{Dinu-Artetxe} dataset are shown in Table \ref{table:prec_all_dinu}. In Table \ref{table:prec_all_muse} and Table \ref{table:prec_all_dinu}, $\rightarrow$' refers to the direction from English to other languages and  '$\leftarrow$' refers to the direction from other languages to English.

\begin{table*}[!h]
\centering
% \vspace{-0.15cm}
% \setlength{\abovecaptionskip}{0.2cm}
% \setlength{\belowcaptionskip}{-0.4cm}
% \resizebox{0.45\textwidth}{!}{
\begin{tabular}{c|c|ccc}
\hline
\hline
language  & \multirow{2}{*}{models} & \multicolumn{3}{c}{$\rightarrow$}\\
pairs & & P@1 & P@5 & P@10 \\
\hline 
\multirow{4}{*}{de} 
& MUSE & 65.57 & 79.61 & 83.23 \\
& VecMap & 65.73 & 79.73 & \textbf{83.37} \\
& \textbf{AutoEnc} & 65.76 & 79.85 & 82.98  \\
& \text{Ours}  & \textbf{66.01} & \textbf{80.05} & 83.14 \\
\hline 
\multirow{4}{*}{es} 
& MUSE & 67.89 & 81.74 & 84.73 \\
& VecMap & 69.31 & 82.11 & 85.33 \\
& \textbf{AutoEnc} & 69.25 & 82.46 & 85.25  \\
& \text{Ours}  & \textbf{69.53} & \textbf{82.66} & \textbf{85.51} \\
\hline
\multirow{4}{*}{it} 
& MUSE & 67.45 & 81.83 & 85.69  \\
& VecMap & 68.71 & \textbf{83.60} & 86.04 \\
& \textbf{AutoEnc} & 68.76 & 83.22 & 86.00 \\
& \text{Ours}  & \textbf{68.78} & 83.35 & \textbf{86.06} \\
\hline 
\multirow{4}{*}{fi} 
& MUSE & 30.76 & 49.57 & 57.47  \\
& VecMap & 37.95 & 58.22 & 65.23 \\
& \textbf{AutoEnc} & 36.38 & 57.60 & 64.24  \\
& \text{Ours}  & \textbf{38.35} & \textbf{58.56} & \textbf{65.25} \\
\hline\hline
\end{tabular}
% }
\caption{BLI translation precision of \textbf{Dinu-Artetxe}. The highest precision values are boldfaced. "-" indicates the value is not available}
\label{table:prec_all_dinu}
\end{table*}

\begin{table*}[!h]
\centering
% \small
\vspace{-1.5cm}
% \setlength{\abovecaptionskip}{0.2cm}
% \setlength{\belowcaptionskip}{-0.4cm}
% \resizebox{0.45\textwidth}{!}{
\begin{tabular}{c|c|cccccc}
\hline
\hline
language  & \multirow{2}{*}{models} & \multicolumn{3}{c}{$\rightarrow$} & \multicolumn{3}{c}{$\leftarrow$} \\
pairs & & P@1 & P@5 & P@10 & P@1 & P@5 & P@10 \\
\hline 
\multirow{4}{*}{de} 
& MUSE & 74.70 & 88.80 & 91.56 & 72.49 & 85.45 & 88.36 \\
& VecMap & 75.33 & \textbf{89.66} & \textbf{92.26} & 74.00 & 85.73 & 88.53 \\
& \textbf{AutoEnc} & 75.72 & 89.50 & 92.18 & 74.35 & 85.67 & 88.77 \\
& \text{Ours}  & \textbf{75.79} & 89.51 & 91.78 & \textbf{74.75} & \textbf{85.89} & \textbf{88.96} \\
\hline 
\multirow{4}{*}{fr} 
& MUSE & 82.21 & 90.53 & 92.78 & 82.36 & 91.34 & 93.08 \\
& VecMap & 82.20 & 91.13 & \textbf{93.20} & \textbf{83.73} & \textbf{91.80} & 93.33 \\
& \textbf{AutoEnc} & \textbf{82.55} & \textbf{91.27} & 93.03 & 83.33 & 91.34 & 93.38 \\
& \text{Ours}  & 82.53 & 91.17 & 93.05 & 83.51 & 91.47 & \textbf{93.41} \\
\hline 
\multirow{4}{*}{ru} 
& MUSE & 40.65 & 68.78 & 75.53 & 58.46 & 75.65 & 79.76 \\
& VecMap & 47.13 & 70.20 & 75.93 & 64.13 & 77.33 & 81.20 \\
& \textbf{AutoEnc} & 47.29 & 73.41 & 78.91 & 64.48 & 77.40 & 81.11 \\
& \text{Ours}  & \textbf{47.55} & \textbf{73.83} & \textbf{79.36} & \textbf{65.08} & \textbf{77.75} & \textbf{81.45} \\
\hline 
\multirow{4}{*}{es} 
& MUSE & 82.36 & 91.17 & 93.60 & 84.02 & 91.90 & 93.54 \\
& VecMap & 82.33 & 91.33 & 92.91 & 84.33 & 92.26 & 93.86 \\
& \textbf{AutoEnc} & 82.25 & 90.70 & \textbf{92.96} & 84.31 & \textbf{92.48} & 94.34 \\
& \text{Ours}  & \textbf{82.36} & \textbf{90.80} & 92.95 & \textbf{84.43} & 92.42 & \textbf{94.37} \\
\hline 
\multirow{4}{*}{it}
& MUSE  & 77.87 & 88.57 & 91.00 & 77.30 & 88.07 & 90.27  \\
& VecMap  & 78.67 & \textbf{89.26} & 91.40 & 79.40 & 88.00 & 90.06  \\
& \textbf{AutoEnc}  & 78.97 & 89.13 & 91.60 & 80.10 & \textbf{88.36} & 90.46  \\					
& \text{Ours}   & \textbf{79.04} & 89.11 & \textbf{91.69} & \textbf{80.31} & 88.29 & \textbf{90.47}  \\
\hline
\hline 

\multirow{4}{*}{he} 
& MUSE & 38.07 & 60.47 & 67.20 & 52.33 & 6q8.09 & 72.16 \\
& VecMap & 41.20 & 62.06 & 69.00 & 52.80 & 66.96 & 72.16 \\
& \textbf{AutoEnc} & 43.71 & 65.39 & 70.36 & 57.76 & 70.21 & 73.50 \\
& \text{Ours}  & \textbf{44.67} & \textbf{65.60} & \textbf{70.60} & \textbf{58.01} & \textbf{70.35} & \textbf{73.71} \\
\hline
\multirow{4}{*}{tr} 
& MUSE & 45.40 & 64.80 & 72.16 & 59.65 & 73.93 & 78.21 \\
& VecMap & 48.06 & 66.33 & 72.00 & 59.64 & 72.51 & 75.78 \\
& \textbf{AutoEnc} & 49.31 & 69.33 & 74.79 & 61.01 & 73.36 & 76.37 \\
& \text{Ours}  & \textbf{50.45} & \textbf{70.50} & \textbf{75.73} & \textbf{61.19} & \textbf{74.08} & \textbf{77.18} \\
\hline 
\multirow{4}{*}{bg} 
& MUSE & 41.97 & 63.75 & 70.49 & 56.29 & 72.17 & 76.74 \\
& VecMap & 44.46 & 64.13 & 68.87 & 57.80 & 71.33 & 74.93 \\
& \textbf{AutoEnc} & 46.62 & 67.63 & 73.85 & \textbf{61.01} & \textbf{75.03} & \textbf{78.76} \\
& \text{Ours}  & \textbf{46.77} & \textbf{68.60} & \textbf{74.53} & 60.85 & 74.42 & 78.17 \\
\hline 
\multirow{4}{*}{hi} 
& MUSE & 30.40 & 47.07 & 52.64 & 39.55 & 57.97 & 63.13 \\
& VecMap & 35.33 & 49.86 & 53.80 & 37.57 & 52.17 & 55.86 \\
& \textbf{AutoEnc} & 35.74 & 50.68 & 56.57 & 46.58 & 62.49 & 66.63 \\
& \text{Ours}  & \textbf{36.88} & \textbf{52.69} & \textbf{58.62} & \textbf{49.27} & \textbf{65.82} & \textbf{70.10} \\
\hline 
\multirow{4}{*}{fi} 
& MUSE  & 42.70 & 67.13 & 74.10 & 59.30 & 74.20 & 78.40  \\
& VecMap  & 46.13 & 70.26 & 76.33 & 62.27 & 76.46 & 80.20  \\
& \textbf{AutoEnc}  & 44.65 & 63.05 & 74.33 & 62.59 & 75.87 & 79.43  \\								
& \text{Ours} & \textbf{48.28} & \textbf{72.11} & \textbf{78.20} & \textbf{64.66} & \textbf{77.16} & \textbf{80.87}  \\
\hline
\hline 

\multirow{4}{*}{ar} 
& MUSE & 30.94 & 50.34 & 58.20 & 44.48 & 60.94 & 65.91 \\
& VecMap & 34.79 & 55.80 & 62.60 & 49.26 & 65.06 & 68.14 \\
& \textbf{AutoEnc} & 36.28 & 56.84 & 64.15 & 53.28 & 67.65 & 71.24\\
& \text{Ours}  & \textbf{36.79} & \textbf{58.27} & \textbf{66.17} & \textbf{54.42} & \textbf{69.31} & \textbf{72.64}\\
\hline 
\multirow{4}{*}{ms} 
& VecMap  & 50.27 & 64.40 & 69.06 & 47.10 & 61.64 & 65.04  \\
& \textbf{AutoEnc} & \textbf{53.42} & \textbf{68.66} & \textbf{73.95} & 48.53 & 62.61 & 66.77  \\
& \text{Ours}   & 52.64 & 67.91 & 72.98 & \textbf{51.22} & \textbf{66.12} & \textbf{70.32}  \\
\hline 
\multirow{2}{*}{ja} & VecMap & 37.56  & 53.39 & 59.08 & 26.12 & 35.29 & 38.87 \\
% & MUSE  & - & - & - & - & - & -  \\
% & \textbf{AutoEnc}  & - & - & - & - & - & -  \\
& \text{Ours}   & \textbf{46.67} & \textbf{62.59} & \textbf{67.15} & \textbf{31.11} & \textbf{44.87} & \textbf{49.19}  \\
\hline 
\multirow{1}{*}{zh} 
% & VecMap & 0.06 & 0.13 & 0.20 & 0.06 & 0.07 & 0.20 \\
% & MUSE  & - & - & - & - & - & -  \\
% & \textbf{AutoEnc}  & - & - & - & - & - & -  \\
& \text{Ours}   & \textbf{41.45} & \textbf{59.75} & \textbf{65.24} & \textbf{32.62} & \textbf{51.53} & \textbf{57.89}  \\

\hline\hline
\end{tabular}
% }
\caption{BLI translation precision of \citet{conneau2017word}. The highest precision values are boldfaced and precision for failed runs are skipped.}
\label{table:prec_all_muse}
\end{table*}

\section{Unsupervised Validation Criterion}
\label{valid_crit}
We apply the unsupervised validation criterion proposed by \citet{conneau2017word} to select the best model. Specifically, we consider 10,000 most frequent source words and find their nearest neighbors in the target space via the current mapping. We maintain a pseudo dictionary that consists of the words with their translations if the translations are also among the 10,000 most frequent words in the target language space. Then among these pseudo translation pairs, the average cosine similarity is calculated as the validation criterion, i.e., we save the model with the largest average cosine similarity. This average cosine value is empirically highly correlated with the mapping quality \cite{conneau2017word}.

\section{Related Work on CLWE and Multilingual Pre-trained Language Models}
\label{clwe_mplm}
Multilingual pre-trained language models (MPLMs) have shown impressing results on cross-lingual tasks by pre-training a single model to handle multiple languages (\citealp{devlin2018bert}; \citealp{lample2019cross}; \citealp{conneau2019unsupervised}; \citealp{ouyang2020ernie}). The shared information among multiple languages is implicitly explored by the overlapped subword vocabulary. Recent works have shown that CLWE is effective to improve the cross-lingual transferability of MPLMs. \citet{chronopoulou2021improving} utilize the mapped CLWEs to initialize the embedding layer of multilingual pre-trained language models and achieve a little improvement on the unsupervised machine translation task. \citet{vernikos2021subword} transfer a pretrained Language Model from one language ($L_1$) to another language ($L_2$) by initializing the embedding layer in $L_2$ as the embeddings of aligned words in $L_1$. They observe performance improvement in the zero-shot XNLI task and the machine translation task. The research of CLWE may benefit recent MPLMs and we leave it for future work.

\end{document}